%% file: icpr_6pages.tex
\documentclass[10,a4paper,conference]{IEEEtran}
\ifCLASSINFOpdf
\else
\fi

\usepackage{times}  
\usepackage{helvet}  
\usepackage{courier}  
\usepackage{url}  
\usepackage{graphicx}  

\usepackage{amsmath}
\usepackage{booktabs} 
\usepackage{tabularx}
\usepackage{multirow}

\usepackage{algorithm}
\usepackage{algorithmic}
\usepackage[table,xcdraw]{xcolor}

\usepackage{tikz}
\usetikzlibrary{shapes}
\usetikzlibrary{arrows}
\usetikzlibrary{positioning}

\begin{document}
%

\title{An Approximate Bayesian Long Short-Term Memory Algorithm for Outlier Detection}



%
\author{\IEEEauthorblockN{Chao Chen\IEEEauthorrefmark{1},
Xiao Lin\IEEEauthorrefmark{2} and
Gabriel Terejanu\IEEEauthorrefmark{3}}

\IEEEauthorblockA{Computer Science and Engineering Department, \\
University of South Carolina, Columbia, South Carolina\\ }

\IEEEauthorblockA{\IEEEauthorrefmark{1}chen288@email.sc.edu,
\IEEEauthorrefmark{2} lin65@email.sc.edu,
\IEEEauthorrefmark{3} terejanu@cec.sc.edu}
}

%


\maketitle

\begin{abstract}

Long Short-Term Memory networks trained with gradient descent and back-propagation have received great success in various applications. However, point estimation of the weights of the networks is prone to over-fitting problems and lacks important uncertainty information associated with the estimation. However, exact Bayesian neural network methods are intractable and non-applicable for real-world applications. In this study, we propose an approximate estimation of the weights uncertainty using Ensemble Kalman Filter, which is easily scalable to a large number of weights.
Furthermore, we optimize the covariance of the noise distribution in the ensemble update step using maximum likelihood estimation. To assess the proposed algorithm, we apply it to outlier detection in five real-world events retrieved from the Twitter platform.

\end{abstract}

%
\IEEEpeerreviewmaketitle

\input{section/introduction.tex}
\input{section/methodology.tex}

\input{section/experiment.tex}

	\input{section/discussion.tex}

\IEEEtriggercmd{\enlargethispage{-5in}}



\bibliographystyle{IEEEtran}
\bibliography{reference} 
%

\end{document}

%% file: section/introduction.tex
\section{Introduction}

The recent resurgence of neural network trained with backpropagation has established state-of-art results in a wide range of domains. However, backpropagation-based neural networks (NN) are associated with many disadvantages, including but not limited to, the lack of uncertainty estimation, tendency of overfitting small data, and tuning of many hyper-parameters. In backpropagation NNs, the lack of uncertainty information is due to the weights that are treated as point estimates tuned with gradient-descent methods. By contrast, \textit{Bayesian neural networks (BNN)}~\cite{neal1996bayesian} can cope with some of these problems by assigning a prior distribution on the parameters~\cite{gal2016a}. Nonetheless, the Bayesian inference in BNNs is intractable, and researchers have developed various approximate methods to estimate the uncertainty of the weights.

Blundell et al.~\cite{blundell2015weight} proposed a backpropagation-based algorithm which could learn the posterior distribution of the weights by training an ensemble of networks. Specifically, the method, termed as \textit{Bayes by Backprop}, learned the uncertainty of the weights by minimising the variational free energy on the marginal likelihood.

Additionally, expectation propagation were applied to estimate the uncertainty of NNs. Hern\'{a}ndez-Lobato and Adams~\cite{lobato2015prob} developed a scalable algorithm which propagated probabilities forward through the network to obtain the marginal likelihood and then obtained the gradients of the marginal probability in the backward step. Similarly, Soudry et al.~\cite{soudry2014expectation}, described an expectation propagation algorithm aiming at approximating the posterior of the weights with factorized distribution in an online setting.

Variational inference was also proved to be theoretically equivalent to the dropout method that was widely used as a regularization techniques for deep NNs~\cite{gal2016dropoutas}. Furthermore, the authors developed tools to extract model uncertainty from dropout NNs. However, variational estimation typically underestimate the uncertainty of the posterior because they ignore the posterior correlation between the weights.

Kalman Filter (KF) is a common approach for parameter estimation in linear dynamic systems. There are a number of work conducted to estimate the parameters of NNs with Extended Kalman Filters (EKF)~\cite{rivals1998a}, which perform parameter estimation for non-linear dynamical systems. However, EKFs were criticized for larger errors of the posterior mean and covariance introduced by the first-order linearization of the nonlinear system~\cite{wan2000the}. Instead, Unscented Kalman Filters (UKFs) were explored to estimate the parameters of non-linear models and were claimed to have better performance to capture the higher order accuracy for the posterior mean and covariance~\cite{julier2004unscented}. 

Compared to UKFs, Ensemble Kalman Filters (EnKF) can scale much better with the dimensionality of the state while capturing non-Gaussian distributions. By propagating ensembles rather than mean values and covariances, EnKFs save computation and storage of dealing with large matrices. Thus EnKFs are capable of handing very large state dimensions, which is common in NNs with many weights.

Surpringly, there is very little attention paid to applying EnKFs for parameter estimation in NNs. In an attempt to introduce the EnKFs to the deep learning community, we evaluate the performance of using an EnKF model for the parameter estimation of LSTM and apply it in an outlier detection task. The goal is to model the evolution of the probability distribution of the observed features at time $t$ using Recurrent Neural Networks (RNNs).The probability distribution is then used to determine whether an observation is an outlier.

RNNs are sequence-based networks designed for sequence data. These models have been successfully applied in areas including speech recognition, image caption generation, and machine translation~\cite{sak2014long}. Compared to feed-forward networks, RNNs can capture the information from all previous time steps and share the same parameters across all steps. The term ``recurrent'' means that we can unfold the network, and at each step the hidden layer performs the same task for different inputs. 


Standard RNN is limited by the gradient-vanishing problem. To cope with this issue, Long Short-Term Memory (LSTM)~\cite{hochreiter1997lstm} networks have been developed to maintain the long term dependency by specifying gating mechanism for the passing of information through the hidden layers. Namely, memory blocks replace the traditional hidden units, and store information in the cell variable. There are four components for each memory block, which include a memory cell, an input gate, an output gate, and a forget gate. 

We propose a Bayesian LSTM where the uncertainty in the weights is estimated using EnKF. To mitigate the underestimation of error covariance due to various sources such as model errors, nonlinearity, and limited ensemble size, in this study we optimize the covariance inflation using maximum likelihood estimation. To assess the proposed algorithm, we apply it to outlier detection in five real-world events retrieved from the Twitter platform.

In the following methodology section we introduce the LSTM, the Bayesian inference using the proposed EnKF, and their application to general outlier detection problems. This will be followed by the subevent detection application in Twitter streams, where the problem specifics and numerical results are presented in the experiment section.

%% file: section/methodology.tex
\section{Methodology}

Given an observed sequence of features, $y^*_1 \ldots y^*_t$, the goal is to contruct the predictive probability density function (pdf) $p(y_{t+1}|y_{1:t} = \{y^*_1 \ldots y^*_t\})$ using a Bayesian LSTM. This pdf is then used to determine whether the next observation $y_{t+1}^*$ is an outlier ($*$ denotes the actual observation). 

\subsection{Long Short-Term Memory (LSTM)}

In LSTM each hidden unit in Figure~\ref{rnns} is replaced by a memory cell. Each memory cell is composed of an input gate, a forget gate, an output gate, and an internal state, which process the input data through the gated mechanism depicted in the following formula.  

\begin{align}
i_{t} &= \sigma(W_{ix}x_{t} + W_{im}m_{t-1} + b_{i})\\
f_{t} &= \sigma(W_{fx}x_{t} + W_{mf}m_{t-1} + b_{f})\\
c_{t} &= f_{t} \odot c_{t-1} + i_{t} \odot g(W_{cx}x_{t} + W_{cm}m_{t-1} + b_{c})\\
o_{t} &= \sigma(W_{ox}x_{t} + W_{om}m_{t-1} + b_{o})\\
m_{t} &= o_{t} \odot h(c_t)\\
y_{t} &= W_{ym}m_{t} + b_{y}
\end{align}

Here, $\sigma$ is the logistic sigmoid function, $i$, $f$, $o$, and $c$ represent the three gates and the internal state, $W$ is the weight matrix, $b$ represents the bias term, $m$ is the cell output activation vector, $\odot$ is element-wise product, $g$ and $h$ are $tanh$ activation functions, and $x$ and $y$ represent the input and the output vector, respectively.


\subsection{Ensemble Kalman Filter (EnKF)}
The ensemble Kalman filter (EnKF) is an approximate inference for the Bayesian nonlinear filtering problem. It can deal with extremely high-dimensional and nonlinear applications~\cite{evensen2003the}. In EnKF, the probability distribution of state variables is described by ensemble members. Each ensemble member is updated in a similar way as in the Kalman filter. Consider the following system
\begin{align}
u_{k} &= F(u_{k-1})  \label{eq:h}\\
d_{k} &= H u_{k} + \epsilon \label{eq:v} 
\end{align}
where $u$ is the state variable and $d$ is the measurement perturbed by the noise $\epsilon$.
\begin{equation}
\epsilon \sim \mathcal{N}(0, \sigma_{\epsilon}^2)~. \nonumber
\end{equation}
Here we use $\lbrace u^{j}_{k-1}\rbrace_{j=1 ... N}$ to denotes ensemble members of $u_{k-1}$. By propagating them through Eq.~\eqref{eq:h}, we can get predictions $\lbrace u^{j}_{k|k-1}\rbrace_{j=1 ... N}$ of $u_{k}$. Once measurements $d_k$ is obtained,  ensemble members of $u_{k}$ can be updated as follows:
\begin{align}
 &u^{j}_k = u^{j}_{k|k-1} + \Sigma_{k|k-1}H^{T} \times \nonumber \label{update}\\
& \quad \quad \quad \quad [H\Sigma_{k|k-1}H^{T} + R_{e}]^{-1}[d^{j}_{k} - Hu^{j}_{k|k-1}] \\
 &\Sigma_{k|k-1} = \overline{[u_{k|k-1} - \overline{u}_{k|k-1}][u_{k|k-1} - \overline{u}_{k|k-1}]^{T}}\\
& \Sigma_{k} = \overline{[u_{k} - \overline{u}_{k}][u_{k} - \overline{u}_{k}]^{T}}
\end{align}
where the perturbed measurements $d^{j}_{k} = d_{k} + \epsilon^{j}$ and $\epsilon^{j}$ is a sample of $\epsilon$. $R_{e} = \overline{\epsilon\epsilon^{T}}$ is the sample covariance matrix of $\epsilon$. Using these perturbed measurements one can guarantee the same results as Kalman filter when the ensemble size is infinite \cite{evensen2003the}.

\subsection{Bayesian LSTM using EnKF}

A RNN can be represented as 
\begin{equation}	
y_k = f(x_k, w) + \epsilon, \quad k = 1,2,...,M \label{ml}
\end{equation}
where $(x_k, y_k)$ is the training data, $w$ is the parameter vector (weights), $f$ is a nonlinear neural network mapping (e.g. LSTM), and $\epsilon$ is the noise which compensates for the difference between outputs of neural network and real target values. 

Let $D$ indicate the training data,  $D = \lbrace (x_k,y_k) \rbrace_{k=1,2,...,M}$. Given a new input $x^*$, we are interested in the predictive distribution 
\begin{equation}  
p(y^*|D, x^*) = \int p(y^*|x^*,w)p(w|D)dw~. \label{integral}
\end{equation}
Here $p(w|D)$ is the conditional distribution of $w$ given the training data $D$, which can be obtained via Bayes' rule:
\begin{equation}
p(w|D) = \frac{p(D|w)p(w)}{p(D)}. \label{bayes}
\end{equation}
$p(w)$ is the prior distribution of the weights $w$, $p(D)$ is the evidence, and $p(D|w)$ is the likelihood which can be obtained though Eq.~\eqref{ml}. 

To evaluate the integral in Eq.~\eqref{integral}, we need to find a solution for Eq.~\eqref{bayes}. Since the neural network is a nonlinear mapping function, a common way to solve Eq.~\eqref{bayes} is using Monte Carlo methods. Suppose $N$ samples $\lbrace w_j\rbrace_{j=1,...,N}$ of $p(w|D)$ are available, and $\delta(\cdot)$ represents the dirac function. Then $p(y^*|D, x^*)$ can be obtained as follows:
\begin{align}
p(y^*|D, x^*) &= \frac{1}{N}\sum_{j=1}^{N}p(y^*|x^*,w_j)\\ \label{monteCarlo}
&= \frac{1}{N}\sum_{j=1}^{N}\delta (y^* - f(x^*,w_j))
\end{align}
Let $\lbrace y^*_j \rbrace_{j=1,...,N}$ denotes samples of $p(y^*|D, x^*)$, then we have $y^*_j = f(x^*,w_j)$

In this paper, we use EnKF to estimate the uncertainty of the weights $w$. The corresponding dynamic system is shown in Eq.~\eqref{wk} and Eq.~\eqref{update}.
\begin{align}
w_k &= w_{k-1} \label{wk}\\
y_k &= f(x_k, w_k) + \epsilon \label{update}
\end{align}
$w$ has a prior distribution $p(w) = \mathcal{N}(w; 0, \sigma^2_w I_p)$ and $\epsilon$ is a white noise with distribution $p(\epsilon) = \mathcal{N}(\epsilon; 0, \sigma^2_{\epsilon}I_q)$. Here $p$ and $q$ represent the dimensionality of features and targets, respectively. 

In order to preserve the relation between consecutive data,  training data are sent to LSTM in batch. Suppose the batch size is $s$ and the number of weights is $l$. Since weights $w$ is the quantity that needs to be estimated, we augment the output of neural network with $w$ to form an augmented state variable $U_k = [F_k, w]$. Here $F_k$ includes all the outputs of the $k$th batch $\lbrace f(x_{k,i}, w) \rbrace_{i=1,...,s}$. The matching measurement model of Eq.~\eqref{eq:v} is given by Eq.~\eqref{mea_model}.
\begin{align}
Y_k &= H U_k + \epsilon   \label{mea_model}\\
\epsilon &\sim \mathcal{N}(0, \sigma^2_{\epsilon}I_{sq}) \nonumber\\
H &= [I_{sq}, 0_{sq\times l}] \nonumber \\
Y_k &= [y_{k,1}^{T},y_{k,2}^{T},...,y_{k,s}^{T}]^{T} \nonumber\\
U_k &= [f(x_{k,1},w)^{T},f(x_{k,2},w)^{T},...,f(x_{k,s},w)^{T}, w_1,w_2,...w_l]^{T} \nonumber
\end{align}

Before inference, two hyperparameters $\sigma^2_w$ and $\sigma^2_{\epsilon}$ need to be determined. A common way is to maximize evidence $p(D|\sigma^2_w, \sigma^2_{\epsilon})$ with respect to $\sigma^2_w$ and $\sigma^2_{\epsilon}$. 
\begin{equation}
p(D|\sigma^2_w, \sigma^2_{\epsilon}) = \int p(D|w, \sigma^2_{\epsilon})p(w|\sigma^2_w) \textit{d}w
\end{equation}
This has been successfully applied to Bayesian linear regression. However, for nonlinear models, it is difficult to evaluate the integral above. Here, we fix $\sigma^2_w$ and estimate $\sigma^2_{\epsilon}$ by maximizing $p(D|\sigma^2_{\epsilon})$. 
Under the assumption that the data points are generated independently, we have
\begin{equation}
p(D|\sigma^2_{\epsilon}) = \prod^{M}_{j=1}p(y_j|x_j,\sigma^2_{\epsilon}). 
\end{equation}
%


The log-evidence is given by
\begin{align}
\textrm{ln}p(D|\sigma^2_{\epsilon}) &= \textrm{ln}\prod^{M}_{j=1}p(y_j|\sigma^2_{\epsilon}) \nonumber \\
&= \sum^{M}_{j=1} \textrm{ln}p(y_j|\sigma^2_{\epsilon}) \nonumber \\
&= \sum^{M}_{j=1}\textrm{ln} \textrm{E}\bigg[\mathcal{N}(y_j; f_i, \sigma^2_{\epsilon}I_q)\bigg] \nonumber
\end{align}

Here $f_i = f(x,w_i)$. Since $\textrm{log}$ is a concave function, according to Jensen's inequality, we have

%
\begin{align}
\textrm{ln}p(D|\sigma^2_{\epsilon}) \geq& \sum^{M}_{j=1}\textrm{E}\bigg[\textrm{ln}\mathcal{N}(y_j; f_i, \sigma^2_{\epsilon}I_q)\bigg] \label{lower_bound} \\
=&  \frac{1}{N}\sum^{M}_{j=1}\sum^{N}_{i=1}\textrm{ln}\mathcal{N}(y_j; f_i, \sigma^2_{\epsilon}I_q) \nonumber \\
=& -\frac{qM}{2}\textrm{ln}2\pi - \frac{qM}{2}\textrm{ln}\sigma^2_{\epsilon} \nonumber \\
&- \frac{1}{2N\sigma^2_{\epsilon}}\sum^{M}_{j=1}\sum^{N}_{i=1}(y_j - f_i)^{T}(y_j - f_i) \nonumber
\end{align}


Maximizing the lower bound of log-evidence with respect to $\sigma^2_{\epsilon}$ we obtain
\begin{equation}
\sigma^2_{\epsilon} = \frac{1}{qMN} \sum^{M}_{j=1}\sum^{N}_{i=1}(y_j - f_i)^{T}(y_j - f_i).  \label{sigma_estimate}
\end{equation}

Once the variance of the noise is determined, the EnKF algorithm presented in the previous step is applied to obtain samples/ensembles from the posterior distribution of the weights. 

%
%

\subsection{Outlier Detection}

The inferred distribution of the weights induces a predictive distribution for the next observable $p(y_{t+1}|y_{1:t} = \{y^*_1 \ldots y^*_t\})$. We can use this probability distribution to label the actual observation $y^*_{t+1}$ as outlier. Since each observation is a multi-dimensional feature with dimension $q$, we can use the chi-squared test of the squared Mahalanobis distance~\cite{warren2011use}. The main idea is to identify when a data point falls outside of the multidimensional uncertainty even when the marginal uncertainties capture the observational data.  

The Mahalanobis distance between the actual observation $y^*_{t+1}$ and the predicted uncertainty approximated using a Gaussian distribution $p(y_{t+1}|y_{1:t}) \approx \mathcal{N}(y_{t+1}; \mu_{t+1}, \Sigma_{t+1})$ is given by
\begin{equation}
m_d = \sqrt{(y^*_{t+1}-\mu_{t+1})^T \Sigma_{t+1}^{-1} (y_{t+1}-\mu_{t+1})}~,
\end{equation}
where the sample mean and covariance matrix are obtained using the propagated ensemble memebers to the observable. 

When the square of the Mahalanobis distance passes the following test, the observations is considered to not be an outlier and a plausible outcome of the model. Here the degree of freedom used to obtain $\chi^2_{0.05}$ is $q$. 
\begin{equation}
m_d^2 \le \chi^2_{0.05}
\end{equation}



%% file: section/experiment.tex
\section{Experiments}

\subsection{Subevent Detection}

An event is confined by space and time. Specifically, it consists of a set of sub-events, depicting different facets of an event. As an event evolves, users usually post new statuses to capture new states as sub-events of the main issue. Within an event, some unexpected situations or results may occur and surprise users, such as the bombing during the Boston Marathon and the power outage during the 2013 Superbowl. Sub-event detection provides a deeper understanding of the threats to better manage the situation within a crisis~\cite{pohl2013}.

By formalizing it as an outlier detection task, we built dynamic models to detect sub-events based upon the retrieved Twitter data and the proposed window embedding representation described in the following sections.

\subsection{Data}

\begin{table*}[t]
	\caption{Basic information of the five events.}
	\resizebox{\textwidth}{!}{%
		\centering
		\begin{tabular}{|r|c|c|c|l|}
			\hline 
			Event & Collection Starting Time & Event Time & Collection Ending Time & Key Words/Hashtags\\ 
			\hline 
			2013 Boston Marathon & 04/12/2013 00:00:00& 04/15/2013 14:49:00& 04/18/2013 23:59:59& Marathon, \#marathon\\ 
			\hline 
			2013 Superbowl & 01/31/2013 00:00:00& 02/03/2013 20:38:00& 02/06/2014 23:59:59& Superbowl, giants, ravens, harbaugh\\ 
			\hline 
			2013 OSCAR & 02/21/2013 00:00:00& 02/24/2013 20:30:00& 02/27/2013 23:59:59 & Oscar, \#sethmacfarlane, \#academyawards\\ 
			\hline 
			2013 NBA AllStar & 02/14/2013 00:00:0& 02/17/2013 20:30:00 & 02/20/2013 23:59:59 & allstar, all-star\\ 
			\hline 
			Zimmerman Trial & 07/12/2013 11:30:00 & 07/13/2013
			22:00:00 & 07/15/2013 11:30:00 & trayvon, zimmerman\\ 
			\hline 
		\end{tabular}%
		\label{table:summary}
	}
\end{table*}

We collected the data from Jan. 2, 2013 to Oct. 7, 2014 with the Twitter streaming API and selected five national events for the outlier detection task. The five events include the 2013 Boston Marathon event, the 2013 Superbowl event, the 2013 OSCAR event, the 2013 NBA AllStar event, and the Zimmerman trial event. Each of these events consists of a variety of subevents, such as the bombing for the marathon event, the power outrage for the Superbowl event, the nomination moment of the best picture award, the ceremony for the NBA AllStar MVP, and the verdict of the jury for the Zimmerman trial event.

For these case studies, we filtered out relevant tweets with event-related keywords and hashtags, preprocessed the data to remove urls and mentioned users. The basic information of each event is provided in Table~\ref{table:summary}. 

\subsection{Window Embedding}

In computational linguistics, distributed representations of words have shown some advantages over raw co-occurrence count since they can capture the contextual information of words. In particular, GloVe~\cite{pennington2014glove} word representation can capture both the patterns and statistics information of the words, and it is thus successfully applied in many NLP applications. Through some experiments, we decided to use the 100 dimension GloVe vector representation that were trained with 27 billion tweets. We further used the Probabilistic PCA to reduce the vector dimensionality into $d$ latent components that could capture at least 99\% variability of the original information.  

Here, we define sentence embedding as the average of its word vectors. Given a sentence, it consists of n words represented by vectors ${e_1^d, e_2^d, ..., e_n^d}$, and the sentence embedding $s_i^d$ is defined as $\sum_{i=1}^{n}e_i^d/n$. Furthermore, we define a window embedding $w_t^d$ as the average of its sentence vectors. For a given time window, it is composed of $m$ sentence vectors ${s_1^d, s_2^d, ..., s_m^d}$, and a window embedding $w_t^d$ is defined as $\sum_{i=1}^m s_i^d/m$. As we use a moving window approach, we grouped every $l$-size window ${w_1^d, w_2^d, ..., w_l^d}$ into a training input $X$, and label $w_{l+1}^d$ as the training input $Y$. Based upon the grouped data, we can train our proposed multivariate EnKF-LSTM model. With some experiments, we chose $5$ as the number of latent components $d$, $5$ minutes as the time window $t$, and $32$ as the grouping size $l$.

\subsection{Implementation}

\begin{figure}[h]
\centering
   \includegraphics[scale=0.4]{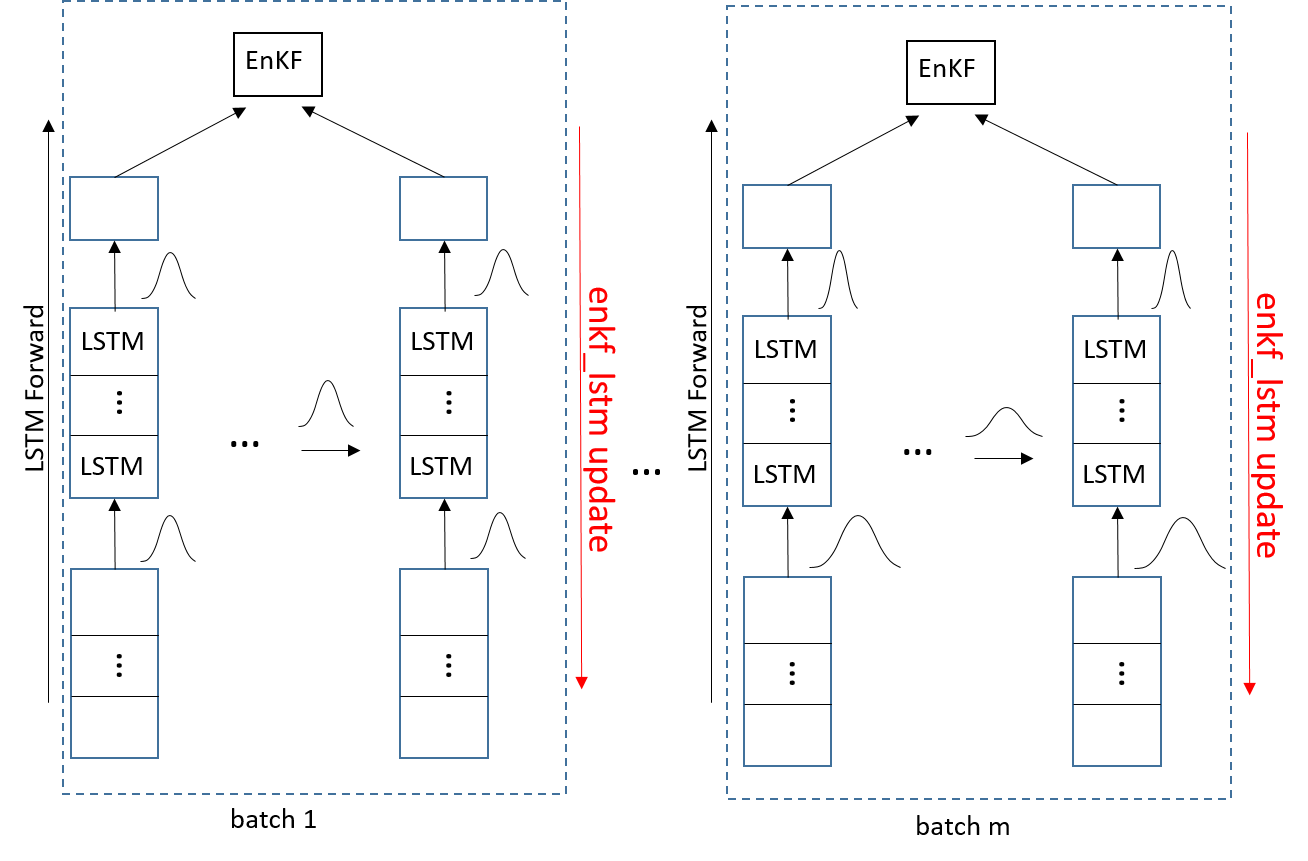}
\caption{Architecture of the network used in this study.} 
\label{enkf_lstm_arch}
\end{figure}

The implemented network architecture is shown in Figure~\ref{enkf_lstm_arch}. The input layer consists 5 nodes, the hidden layer consists 32 LSTM cells, and the output layer consists 5 output nodes. In this implementation, we include the forget gate proposed by~\cite{gers1999learning}. The implementation is based upon Tensorflow,and it could be easily extended for deep architectures or variants of LSTMs. 

Figure~\ref{enkf_lstm_arch} provides an intuitive introduction of the architecture and the proposed algorithm. The algorithm proceeds in a batch mode. At the very beginning, the prior weights are drawn from a multivariate Gaussian distribution. Subsequently, we forward propagate each batch through the LSTM cells, and obtain the network outputs. In terms of the network outputs, we augment them with prior weights and update the augmented variable using EnKF. Then we return the updated weights for the next batch process.

%% file: section/discussion.tex
\section{Results}

\begin{figure*}[!]
\centering
   \includegraphics[width=1.0\textwidth]{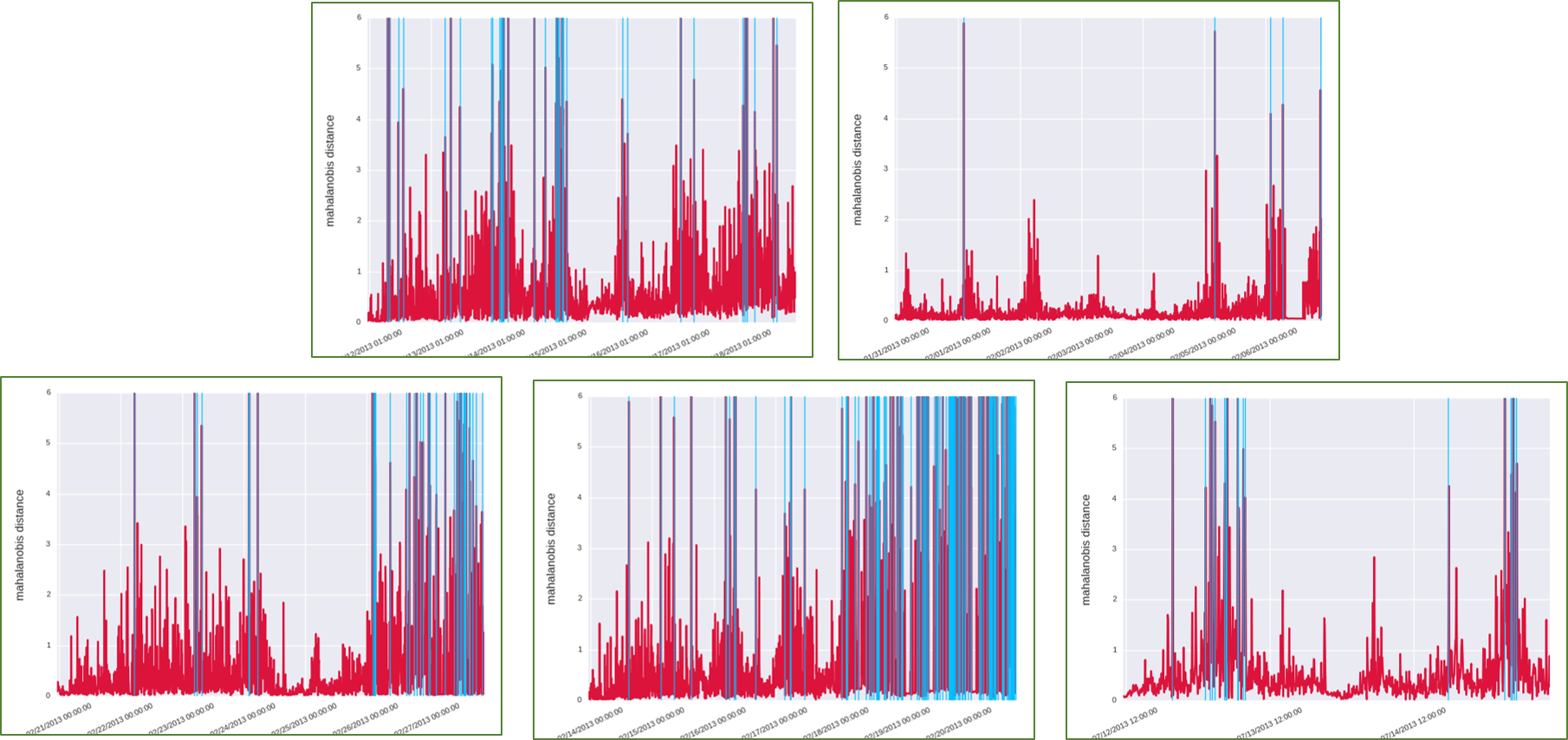}
\caption{Predicted sub-events with the proposed algorithm for the 2013 Boston marathon event, the 2013 Superbowl event, the 2013 OSCAR event, the 2013 NBA AllStar event, and the Zimmerman trial event, respectively (read from top to bottom and left to right). The red color indicates the Mahalanobis distance and the cyan color indicates the identified outliers.} 
\label{outliers_md}
\end{figure*}

The outlier detection results are provided in Figure~\ref{outliers_md}. In terms of the results, we observe 37, 5, 39, 131, and 19 identified sub-events, respectively. Of those sub-events, 16, 3, 16, 42, and 17 are verified as true sub-events. We set the initial sigma value of the noise covariance matrix in the EnKF update step to 1.0, and then further optimized them to 2.17, 2.15, 2.16, 2.018, and 0.19 with Maximum Likelihood Estimation.

To further evaluate our model, we compared it with Gaussian Process (GP) and MC dropout~\cite{gal2016dropoutas}. The comparison result is provided in Table~\ref{eval_table}. The GP model yielded the best recall value in three of the five events, indicating that it captured most true sub-events. On the other hand, it also misidentified many normal time windows as sub-events, thus yielding many false positives and low precision. Compared to the GP model, our proposed $enkf\_lstm$ algorithm reliably captured many true sub-events and yield the best precision in these five events. Though, on the other hand, it missed capturing some true sub-events and had worse recall performance in three of the five events. In terms of the F1 score, our proposed algorithm has the best performance in three of the five cases. The MC dropout model, however, has the worst performance for this specific outlier detection task. Since MC dropout is mathematically equivalent to variational inference, which under-estimates the uncertainty, the model mislabels many normal time windows as outliers.

\begin{table}[!]
\centering
\caption{Evaluation metrics on different models.}
\label{eval_table}

\resizebox{0.8\columnwidth}{!}{%
\begin{tabular}{|r|c|c|c|c|}
\hline
\rowcolor[HTML]{FFFFFF} 
Event                                                          & Model      & Precision     & Recall        & F1 Score      \\ \hline
\rowcolor[HTML]{ECF4FF} 
\cellcolor[HTML]{ECF4FF}                                       & GP         & 17.1          & \textbf{45.8} & 24.9          \\ \cline{2-5} 
\rowcolor[HTML]{ECF4FF} 
\cellcolor[HTML]{ECF4FF}                                       & ENKF LSTM  & \textbf{43.2} & 24.6          & \textbf{31.3} \\ \cline{2-5} 
\rowcolor[HTML]{ECF4FF} 
\multirow{-3}{*}{\cellcolor[HTML]{ECF4FF}2013 Boston Marathon} & MC Dropout & 10.9          & 30.8          & 16.1          \\ \hline
\rowcolor[HTML]{FFFFFF} 
\cellcolor[HTML]{FFFFFF}                                       & GP         & 20.4          & \textbf{30.5} & \textbf{24.4} \\ \cline{2-5} 
\rowcolor[HTML]{FFFFFF} 
\cellcolor[HTML]{FFFFFF}                                       & ENKF LSTM  & \textbf{60.0} & 10.7          & 18.2          \\ \cline{2-5} 
\rowcolor[HTML]{FFFFFF} 
\multirow{-3}{*}{\cellcolor[HTML]{FFFFFF}2013 Superbowl}       & MC Dropout & 4.9           & 14.3          & 7.3           \\ \hline
\rowcolor[HTML]{ECF4FF} 
\cellcolor[HTML]{ECF4FF}                                       & GP         & 18.2          & \textbf{37.8} & \textbf{24.6} \\ \cline{2-5} 
\rowcolor[HTML]{ECF4FF} 
\cellcolor[HTML]{ECF4FF}                                       & ENKF LSTM  & \textbf{41.0} & 13.6          & 20.4          \\ \cline{2-5} 
\rowcolor[HTML]{ECF4FF} 
\multirow{-3}{*}{\cellcolor[HTML]{ECF4FF}2013 OSCAR}           & MC Dropout & 8.8           & 34.1          & 14.0          \\ \hline
\rowcolor[HTML]{FFFFFF} 
\cellcolor[HTML]{FFFFFF}                                       & GP         & 18.1          & 55.6          & 27.3          \\ \cline{2-5} 
\rowcolor[HTML]{FFFFFF} 
\cellcolor[HTML]{FFFFFF}                                       & ENKF LSTM  & \textbf{32.1} & \textbf{63.6} & \textbf{42.7} \\ \cline{2-5} 
\rowcolor[HTML]{FFFFFF} 
\multirow{-3}{*}{\cellcolor[HTML]{FFFFFF}2013 NBA AllStar}     & MC Dropout & 16.3          & 45.5          & 24.0          \\ \hline
\rowcolor[HTML]{ECF4FF} 
\cellcolor[HTML]{ECF4FF}                                       & GP         & 25.1          & 65.9          & 36.4          \\ \cline{2-5} 
\rowcolor[HTML]{ECF4FF} 
\cellcolor[HTML]{ECF4FF}                                       & ENKF LSTM  & \textbf{89.5} & \textbf{70.8} & \textbf{79.1} \\ \cline{2-5} 
\rowcolor[HTML]{ECF4FF} 
\multirow{-3}{*}{\cellcolor[HTML]{ECF4FF}Zimmerman Trial}      & MC Dropout & 22.5          & 37.5          & 28.1          \\ \hline
\end{tabular}
}
\end{table}

\begin{figure}[!]
\centering 
\includegraphics[width=0.7\columnwidth]{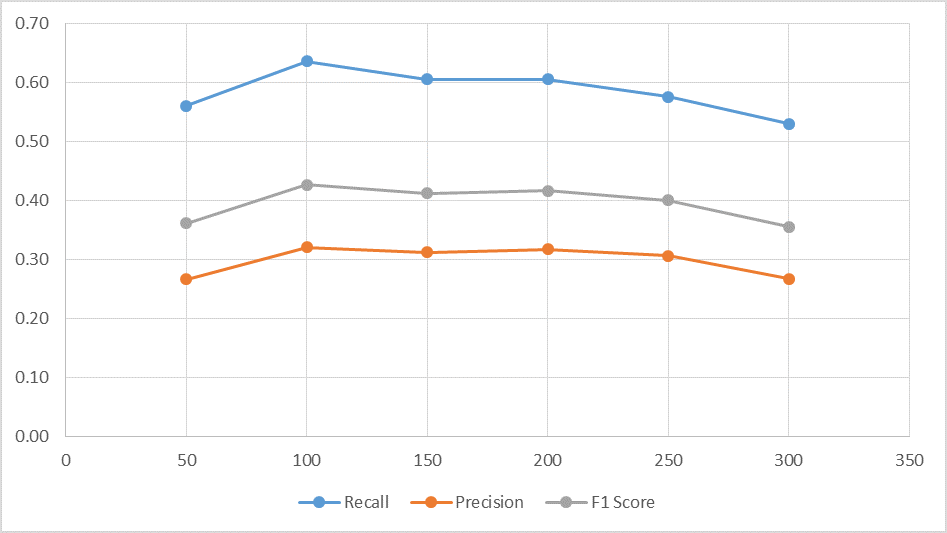}
\label{sensitivity_ensemble}
\caption{Performance of the algorithm on the AllStar event for different ensemble size.}
\end{figure}

\begin{figure}[!]
\centering 
\includegraphics[width=0.7\columnwidth]{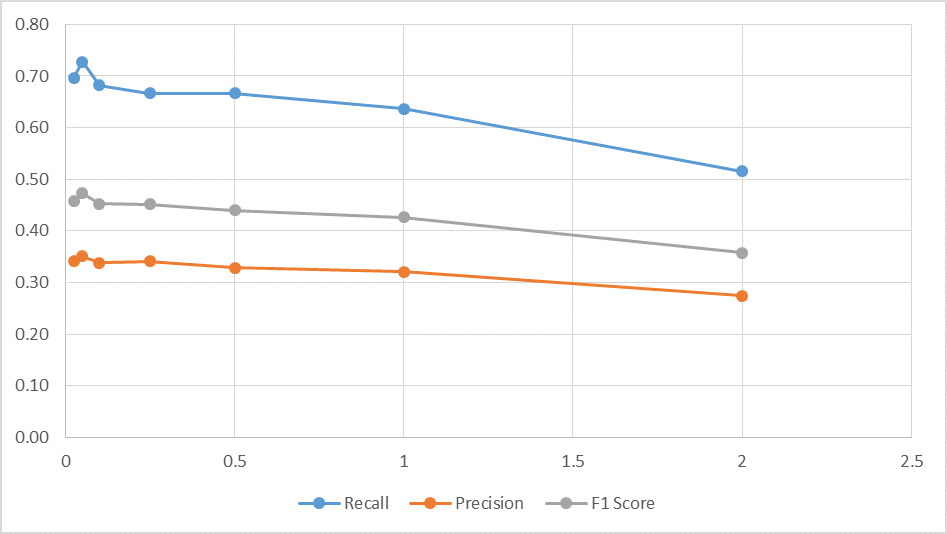}
\label{sensitivity_sigma}
\caption{Performance of the algorithm on the AllStar event for different sigma value.}
\end{figure}

For the proposed algorithm, ensemble size $N$ and the initial sigma value of the noise covariance matrix $\sigma_{\epsilon}^2$ are two important hyper-parameters. To further evaluate their effects on the performance, we provided a sensitivity analysis of the hyper-parameters for the 2013 AllStar event. Based upon Figure~\ref{sensitivity_ensemble}, the algorithm yielded the best result with an ensemble size at 100, and varied slightly with different sizes. According to Figure~\ref{sensitivity_sigma}, the evaluation metrics peaked at 0.05 and then slightly decreased with larger value.

\subsection{Discussion}

In this work, we proposed a novel algorithm to estimate the posterior weights for LSTMs, and we further developed a framework for outlier detection. Based upon the proposed algorithm and framework, we applied them for five real-world outlier detection tasks using Twitter streams. As shown in the above section, the proposed algorithm can capture the uncertainty of the non-linear multivariate distribution. However, the performance of the model is affected by several hyper-parameters, including the number of ensembles, the batch size, the initial sigma value, the number of layers, and the number of nodes in each layer. The performance of the detection is further limited by the choice of window size and word representations. In the future study, we will provide a more detailed analysis of the effects of these hyper-parameters on the model performance and fine-tune them with Bayesian Optimization.